\documentclass[10pt,twocolumn,letterpaper]{article}

\usepackage{cvpr}              
\usepackage[accsupp]{axessibility} 

\usepackage[dvipsnames]{xcolor}

\definecolor{cvprblue}{rgb}{0.21,0.49,0.74}
\usepackage[pagebackref,breaklinks,colorlinks,citecolor=cvprblue]{hyperref}
\usepackage{multirow}
\usepackage{algorithm}
\usepackage{algorithmic}
\usepackage{lipsum}

\def\paperID{9002} 
\def\confName{CVPR}
\def\confYear{2024}

\title{
GoodSAM: Bridging Domain and Capacity Gaps via Segment Anything Model for Distortion-aware Panoramic Semantic Segmentation}

\author{Weiming Zhang$^{1}$ \quad Yexin Liu$^{1}$ \quad Xu Zheng$^{1}$ \quad Lin Wang$^{1}$$^{,2}$\thanks{Corresponding author (e-mail: linwang@ust.hk)}\\
$^{1}$ AI Thrust, HKUST(GZ) \quad $^{2}$Dept. of CSE, HKUST \\
{\tt\small zweiming996@gmail.com, yliu292@connect.hkust-gz.edu.cn, zhengxu128@gmail.com, linwang@ust.hk} \\
\small{Project Page:~\url{https://vlislab22.github.io/GoodSAM/}}
}

\begin{document}

\newcommand\blfootnote[1]{%
\begingroup
\renewcommand\thefootnote{}\footnote{#1}%
\addtocounter{footnote}{-1}%
\endgroup
}

\twocolumn[{
\renewcommand\twocolumn[1][t!]{#1}%
\maketitle
\vspace{-20pt}
\begin{center}
    \centering
    \includegraphics[width=\textwidth]{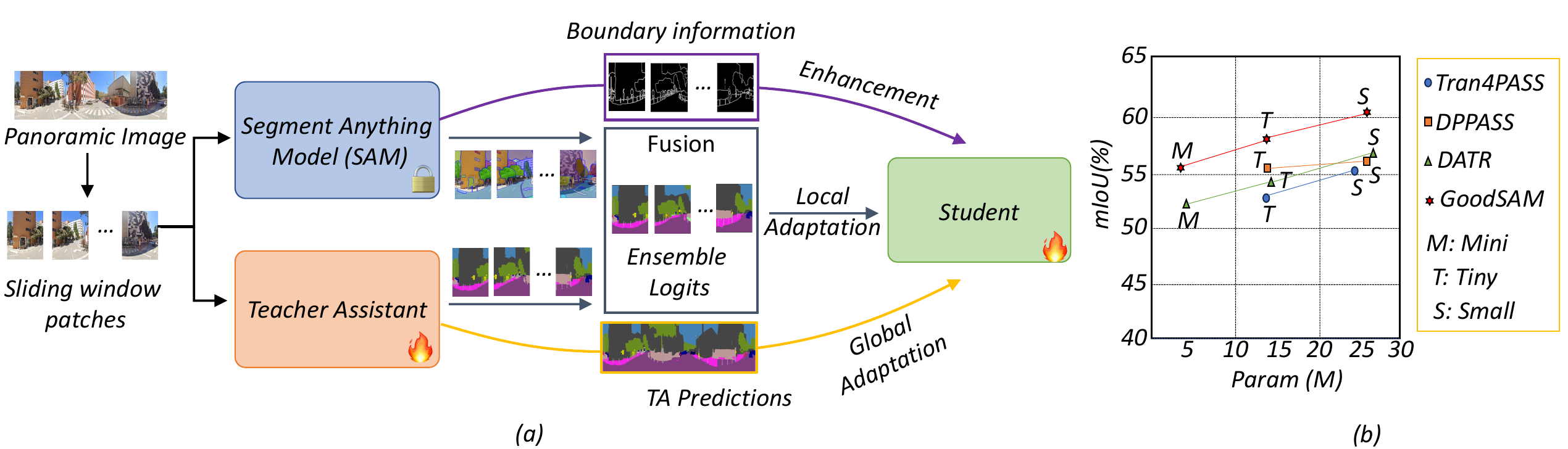}
    \vspace{-20pt}
    \captionof{figure}{(a) Illustration of our GoodSAM, leveraging instance masks and boundary information provided by SAM, coupled with segmentation logits from the teacher assistant, to obtain reliable ensemble logits for knowledge adaptation to our student. (b) Our GoodSAM outperforms SOTA methods ~\cite{zhang2022bending, zheng2023both, zheng2023look} across various model parameter ranges. Notably, GoodSAM-M achieves comparable performance to the SOTA methods with only \textbf{3.7}M parameters.}
    \label{cover figure}
\end{center}}]
\blfootnote{*Corresponding Author}
\begin{abstract}
This paper tackles a novel yet challenging problem: how to transfer knowledge from the emerging Segment Anything Model (SAM) -- which reveals impressive zero-shot instance segmentation capacity -- to learn a compact panoramic semantic segmentation model, \ie, student, without requiring any labeled data. This poses considerable challenges due to SAM's inability to provide semantic labels and the large capacity gap between SAM and the student.
To this end, we propose a novel framework, called \textbf{GoodSAM}, that introduces a teacher assistant (TA) to provide semantic information, integrated with SAM to generate ensemble logits to achieve knowledge transfer.
Specifically, we propose a Distortion-Aware Rectification (DAR) module that first addresses the distortion problem of panoramic images by imposing prediction-level consistency and boundary enhancement.
This subtly enhances TA's prediction capacity on panoramic images.
DAR then incorporates a cross-task complementary fusion block to adaptively merge the predictions of SAM and TA to obtain more reliable ensemble logits.
Moreover, we introduce a Multi-level Knowledge Adaptation (MKA) module to efficiently transfer the multi-level feature knowledge from TA and ensemble logits to learn a compact student model.
Extensive experiments on two benchmarks show that our GoodSAM achieves a remarkable \textbf{+3.75\%} mIoU improvement over the state-of-the-art (SOTA) domain adaptation methods,~\eg, \cite{zheng2023look}. Also, our most lightweight model achieves comparable performance to the SOTA methods with only \textbf{3.7}M parameters.
\end{abstract}

\vspace{-20pt}
\section{Introduction}
The burgeoning demand for omnidirectional scene understanding has stimulated the popularity of $360^\circ$ cameras, making them highly suitable and popular for applications, such as self-driving ~\cite{wang2018self,zheng2023look,wang2020bifuse,jayasuriya2020active}.
This has inspired intensive recent research endeavors~\cite{zheng2023both,li2023sgat4pass,zhang2021transfer} for addressing scene understanding tasks, especially panoramic semantic segmentation.
Generally,
Equirectangular Projection (ERP) is commonly adopted to represent the 360$^\circ$ data in 2D planar representation
\footnote{In this paper, omnidirectional and panoramic images are interchangeably used, and ERP images often indicate panoramic images.} to align spherical data with the existing deep learning pipelines.
However, ERP images often suffer from distortion and deformation problems, induced by the non-uniformly distributed pixels\cite{ai2022deep}.
Also, there is a scarcity of well-annotated datasets, which poses a challenge in training effective models for panoramic segmentation.

Therefore, research has been conducted to transfer knowledge from the labeled pinhole image domain to the unlabeled panoramic image domain via unsupervised domain adaptation (UDA) ~\cite{liu2021pano,zhang2021transfer,zhang2022bending,zheng2023both,zheng2023look}. These methods can be primarily divided into three groups: pseudo labeling~\cite{liu2021pano, zhang2021deeppanocontext}, adversarial training~\cite{ma2021densepass, zheng2023both} and prototypical adaptation~\cite{zhang2022bending, zhang2022behind}. However, they require labeled 2D images for training or adopting multi-branch designs~\cite{yang2020ds,zheng2023both}, thus leading to considerable computational costs.
Recently, foundational models have undergone a significant development~\cite{kirillov2023segment,singh2022flava,nguyen2023lvm}. Large visual models (LVMs), such as the Segment Anything Model (SAM)~\cite{kirillov2023segment} -- trained on very large datasets (over 1 billion masks on 11 million images) have received great attention.
The exceptional zero-shot instance segmentation performance on unseen datasets and tasks makes SAM exceptional in serving as a foundational model for many segmentation tasks~\cite{wu2023medical,ma2023segment,zhang2023segment}.

In this paper, we tackle a novel problem: \textit{how to transfer the instance segmentation knowledge from SAM to learn a more compact panoramic semantic segmentation model (\ie, student) without requiring any labeled data?}
This poses considerable challenges to directly apply SAM for panoramic semantic segmentation
due to two reasons: \textbf{1)} SAM's inability to provide semantic labels, and \textbf{2)} the substantial capacity gap between SAM and the student model. These obstacles render the task of learning a compact student model non-trivial.
To this end, we propose a novel framework, named \textbf{GoodSAM}
to learn a compact student model. Our key insight is to introduce a teacher assistant (TA) to 1) provide semantic labels, enabling to generate the ensemble logits with SAM, and 2) bridge the capacity gap with the student for effective knowledge adaptation.

Our GoodSAM enjoys two key technical contributions.
Specifically, we first propose a Distortion-Aware Rectification (DAR) module to generate reliable ensemble logits (Sec.~\ref{DAR}).
DAR module optimizes ensemble logits from three aspects, see Fig.~\ref{cover figure} (a). Firstly, we leverage an overlapping sliding window strategy~\cite{liu2021swin} to alleviate the adverse effects of the ERP image's large Field of View (FoV) on the performance of both SAM and TA. Secondly, we impose prediction-level consistency for overlapping regions between two adjacent windows and boundary enhancement, using the boundary information provided by SAM, to enhance TA's ability to address the inevitable distortion and object deformation in the ERP images. Finally, we propose a cross-task complementary fusion (CTCF) block, which adaptively combines SAM's instance masks and TA's semantic labels to obtain \textit{high-quality ensemble logits}.
Upon obtaining reliable and distortion-aware ensemble logits, we then introduce the Multi-level Knowledge Adaptation (MKA) module to facilitate the learning of a compact panoramic segmentation student model (Sec.~\ref{MKA}). MKA facilitates multi-level and multi-scale knowledge transfer from TA and ensemble logits, encompassing both whole-image scale and window-based scale to bridge the capacity gap between SAM and the student model and improve the performance of our compact student model.

We conducted extensive experiments to validate our method. As shown in Fig.~\ref{cover figure} (b),  our GoodSAM outperforms SOTA UDA methods across various model parameter ranges. Our GoodSAM's small version, with approximately \textbf{25} million parameters, achieves an impressive \textbf{3.75\%} performance improvement compared to SOTA methods with a similar parameter count. Meanwhile,  our GoodSAM's tiny version achieves comparable performance to the SOTA methods while using only \textbf{3.7} million parameters.

In summary, our contributions are as follows: \textbf{(I)} Our work serves as the \textbf{first} attempt to learn an efficient panoramic semantic segmentation model from SAM. \textbf{(II)} We propose the GoodSAM framework which incorporates DAR and MKA modules to obtain reliable ensemble logits and conduct effective knowledge transfer for panoramic segmentation, respectively. 
\textbf{(III)} We demonstrate the effectiveness of our proposed GoodSAM framework, achieving SOTA performance on panoramic semantic segmentation tasks while maintaining a compact model size.

\vspace{-3pt}
\section{Related Work}
\vspace{-3pt}
\noindent\textbf{Panorama Image Semantic Segmentation.}
The first line of works~\cite{yang2019pass, orhan2022semantic, yang2019can, xu2019semantic, yang2020ds, yang2020omnisupervised} on panoramic semantic segmentation are based on the supervised learning. However, since there is no sufficient panoramic image datasets exist, most of the existing panoramic image semantic segmentation methods are based on unsupervised domain adaptation (UDA)~\cite{zhu2023patch,zhu2023good}. Recent research endeavors have been focused on the UDA for panoramic semantic segmentation approaches, which can be divided into three types, including the pseudo labeling~\cite{liu2021pano, zhang2021deeppanocontext, zhang2017curriculum}, adversarial training~\cite{zhang2021transfer, ma2021densepass, zheng2023both} and prototypical adaptation~\cite{zhang2022bending, zhang2022behind, zheng2024semantics} methods. \textit{Differently, we introduce SAM to the panoramic semantic segmentation task, aiming at transferring the instance segmentation knowledge of SAM to learn a compact student model, assisted by a TA model}.

\begin{figure*}[t]
    \centering
    \includegraphics[width=0.9\textwidth]{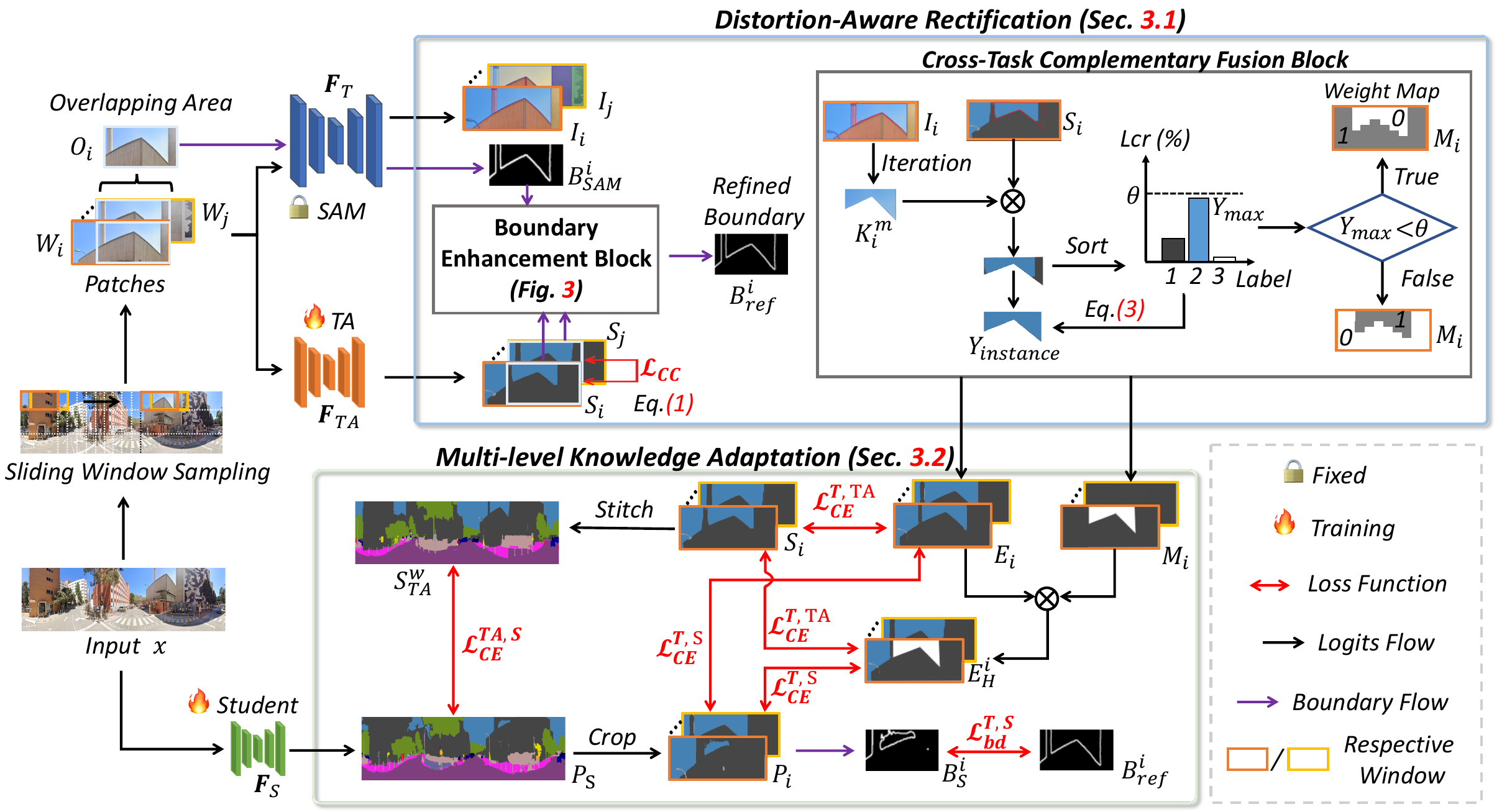}
    \vspace{-8pt}
    \caption{\textbf{Overview of GoodSAM framework}, consisting of three models: SAM, teacher assistant, and student. Our method has two main technical components: the Distortion-Aware Rectification (DAR) module and the Multi-level Knowledge Adaptation (MKA) module.
    }
    \vspace{-15pt}
    \label{framework}
\end{figure*}

\noindent \textbf{Segment Anything Model (SAM)}
It stands as a foundation model for segmentation tasks~\cite{kirillov2023segment}. SAM leverages a dataset of 11 million diverse, high-resolution images, accompanied by 1.1 billion meticulously annotated high-quality segmentation masks for training. This remarkable dataset empowers SAM with robust zero-shot instance segmentation capacity. SAM has been applied to diverse domains, such as medical image segmentation \cite{wu2023medical,ma2023segment,zhang2023segment,roy2023sam}, image editing \cite{yu2023inpaint,zhang2023comprehensive}, and tracking \cite{cheng2023segment,yang2023track}.
However, SAM is limited in providing semantic information, coupled with its prohibitive parameters and domain gap. This poses a challenge for directly achieving panorama semantic segmentation.
To address this, we introduce a TA model that leverages SAM's high-quality instance masks to generate reliable ensemble logits, enabling us to learn a compact yet effective panoramic segmentation model.

\noindent\textbf{Knowledge Adaptation.}
Some methods leverage the expertise of multiple teachers and the similarities between their domains to improve the performance of a student model~\cite{ruder2017knowledge, he2019knowledge, ma2014knowledge}.
The knowledge adaptation lies in identifying the trustworthy knowledge of the teacher's expertise that can be applied effectively in the target domain.
However, existing methods exhibit limitations, such as the risk of the student excessively relying on a teacher's biases, and challenges in generalizing knowledge across tasks or unseen data.
Consequently, directly applying these methods to our panoramic segmentation task is impractical. Intuitively, we propose CTCF block to adaptively combine the outputs from SAM and TA to obtain more reliable ensemble logits. Meanwhile, we propose the MKA module that utilizes TA's outputs and th ensemble logits from the DAR module to perform multi-level and multi-scale knowledge transfer.

\vspace{-5pt}
\section{Methodology}
\noindent\textbf{Overview.}
An overview of our framework is shown in Fig.~\ref{framework}. Given an unlabeled panoramic image $x \in R^{H \times W \times 3}$, we aim to train a compact panoramic semantic segmentation model (\ie, student) $\textbf{F}_S$ under the guidance of SAM $\textbf{F}_T$ (\ie, teacher) and the assistance of the teacher assistant (TA) $\textbf{F}_{TA}$. Note that the TA aims to bridge the capacity gap between SAM and the student during training.
To mitigate the impact of ERP's large FoV, we employ an overlapping sliding window strategy to extract $n$ local patches $\{W_i\}_{i=0}^{n-1}$ from the input ERP images.
Since the distortion in horizontal pixel distribution is more severe than the vertical one, we choose to do horizontal window sliding.
By setting the sliding window movement stride, we can obtain the overlapping area $O_i$ between two horizontally adjacent windows $W_i$ and $W_j$.
Subsequently, patches are input to both SAM $\textbf{F}_T$ and TA $\textbf{F}_{TA}$, yielding their respective predictions $I_i$ and $S_i$. Furthermore, the overlapping area $O_i$ is exclusively input to $\textbf{F}_T$ to derive the corresponding boundary map ($B_{SAM}^i$). For the student model $\textbf{F}_S$, we input the ERP image $x$ to obtain the semantic prediction map $P_S$.
The challenges lie in:
1) effectively fusing the predictions from SAM $\textbf{F}_T$ and TA $\textbf{F}_{TA}$ to obtain more reliable ensemble logits $E_i$ as the supervision for student model $\textbf{F}_S$;
2) effectively performing knowledge adaptation from ensemble logits $E_i$ and $\textbf{F}_{TA}$ to our compact student $\textbf{F}_S$.
To this end, we introduce the GoodSAM framework consisting of two key technical modules: Distortion-Aware Rectification (DAR) Module (Sec.~\ref{DAR}) and Multi-level Knowledge Adaptation (MKA) Module (Sec.~\ref{MKA}). We now describe these modules in detail.

\vspace{-5pt}
\subsection{ Distortion-Aware Rectification (DAR) Module}
\vspace{-2pt}
\label{DAR}
DAR module aims to generate ensemble logits based on SAM and TA. Specifically, as ERP images, in contrast to 2D images, possess a large FoV and distortion problem, impacting the performance of SAM and TA. We introduce the consistency constraint and boundary enhancement.
Moreover, SAM and TA generate different kinds of segmentation maps: SAM provides instance masks, while TA produces semantic maps. Therefore, we propose the cross-task complementary fusion block. We now illustrate the details.

\noindent \textbf{Consistency Constraint.}
It aims to help TA generate distortion-aware semantic maps.
For the adjacent patches $W_i$ and $W_j$, due to distortion of ERP, $\textbf{F}_{TA}$'s predictions $S_i$ and $S_j$ for the overlapping area $O_i$ between $W_i$ and $W_j$ may exhibit discrepancies. Therefore, the purpose of the constraint is to minimize the discrepancies in the overlapping area. For simplicity, we utilize the mean squared error (MSE) loss to ensure the consistency of predictions. As such, we can enhance the $\textbf{F}_{TA}$'s sensitivity to local distortions. Formally, the consistency constraint loss $\mathcal{L}_{CC}$ is:
\begin{equation}
\setlength{\abovedisplayskip}{3pt}
\setlength{\belowdisplayskip}{3pt}
  \mathcal{L}_{CC}=MSE(W_i(O_i),W_j(O_j)),
  \label{eq:W}
\end{equation}
where the $W_i(O_i)$ denotes the TA's prediction within the overlapping area for the $i$ th window.

\begin{figure}[t]
    \centering
    \includegraphics[width=\columnwidth]{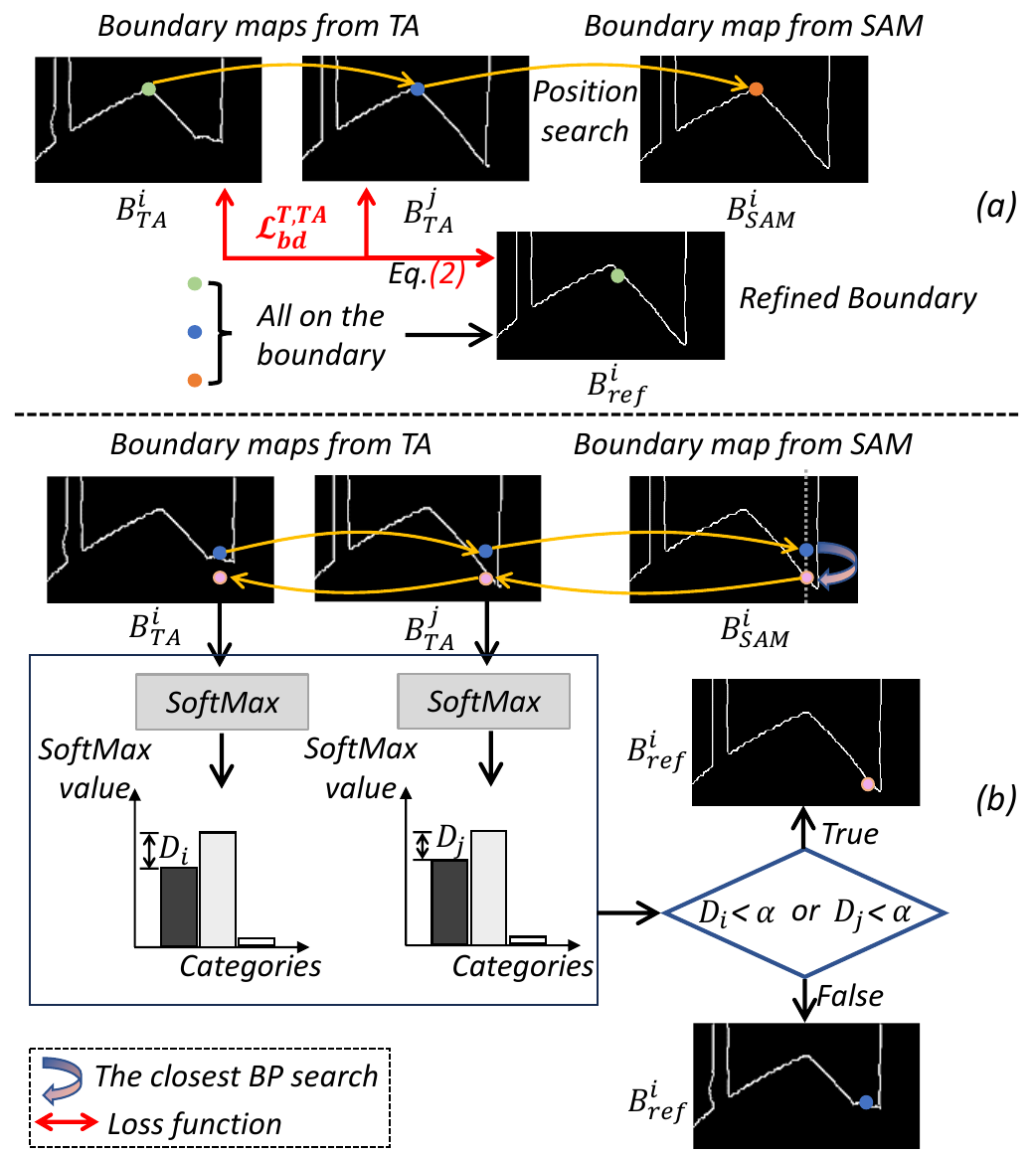}
    \vspace{-18pt}
    \caption{\textbf{Overview of the proposed boundary enhancement block}. In (a), it represents the condition where the pixels at the same positions in all three images are boundary pixels. In (b), it represents the condition where pixels at the same positions are not on the boundary. Additionally, (a) demonstrates the optimization of the boundary enhancement loss for $B_{TA}^i$ and $B_{TA}^j$.
    }
    \vspace{-8pt}
    \label{fig:sec3.2}
\end{figure}

\noindent \textbf{Boundary Enhancement Block.}
Due to SAM's strong zero-shot capability to provide relatively accurate boundary maps, we propose a boundary enhancement block \textit{to refine boundary pixels in TA's predictions}, inspired by \cite{rong2023boundary}.
By improving TA's accuracy in predicting boundary pixels, we aim to alleviate the impact of distortion and object deformation on TA.
This block comprises two components: boundary refinement strategy and boundary-enhanced loss.

As depicted in Fig.~\ref{fig:sec3.2}(a), for two adjacent windows $W_i$ and $W_j$, we obtain two separate boundary maps $B_{TA}^i$ and $B_{TA}^j$ of the overlapping area $O_i$.
The boundary refinement strategy is proposed to identify reliable boundary pixels (denoted as \textcolor{green}{green}, \textcolor{blue}{blue} and \textcolor{orange}{orange} points) within the overlapping area by combining $B_{TA}^i$, $B_{TA}^j$, and $B_{SAM}^i$ of SAM to obtain the refined boundary map $B_{ref}^i$. The detailed Algorithm for this strategy can be found in the \textit{Supplmat}.

Specifically, for the input boundary map $B_{TA}^i$, we first iterate through its boundary pixels and find corresponding pixels in $B_{TA}^j$ and $B_{SAM}^i$ at the same positions.
If the corresponding pixels at the same positions are all on the boundary (See Fig.~\ref{fig:sec3.2} (a)), then the pixel is considered a reliable boundary pixel.
For cases that do not meet the above condition, we find corresponding pixels in $B_{SAM}^i$ at the same positions and identify the nearest boundary pixel (the \textcolor{pink}{pink} point) in the vertical direction (See Fig.~\ref{fig:sec3.2} (b)).

Subsequently, we locate pixels at the same positions (\textcolor{pink}{pink} points) in $B_{TA}^i$ and $B_{TA}^j$, and for each pixel, perform softmax on its logits ($1\times 1 \times C$, where $C$ is the number of categories).
We calculate the difference in the top two softmax values for each corresponding pixel in $B_{TA}^i$ and $B_{TA}^j$, denoting them as $D_i$ and $D_j$ respectively.
When either $D_i$ or $D_j$ has a value less than $\alpha$, we determine that the boundary pixel from $B_{SAM}^i$ exhibits the characteristics of the boundary pixel in the other two boundary maps as well.
Thus, we define the SAM boundary pixel (\textcolor{pink}{pink} point) as reliable.
The parameter $\alpha$ determines the influence strength of SAM boundary pixels. Finally, if none of the above conditions are met, we decide to retain the boundary pixels of $B_{TA}^i$ as reliable pixels.
This way, we attain a refined boundary map $B_{ref}^i$ with all reliable boundary pixels for $O_i$, which is utilized for updating TA $\textbf{F}_{TA}$ and student $\textbf{F}_{S}$.

Next, we introduce a boundary-enhanced loss (See Fig.~\ref{fig:sec3.2} (a)) to encourage TA's boundary pixel predictions to align closely with the refined boundary map:
\begin{small}
\begin{equation}
\setlength{\abovedisplayskip}{3pt}
\setlength{\belowdisplayskip}{3pt}
\begin{aligned}
\mathcal{L}_{bd}^{T,TA}=\sum_{k = 1}^{H\times W}\frac{(|B_{ref}^i-B_{TA}^i|+|B_{ref}^i-B_{TA}^j|)}{C_o},
\label{eq:boundary loss of T and TA}
\end{aligned}
\end{equation}
\end{small}
where $C_o$ denotes the total boundary pixel counts of $B_{ref}$ and $k$ denotes the $k$-th pixel in the boundary map. This explicitly mitigates boundary pixel prediction errors caused by the distortion of ERP.
\vspace{2pt}

\noindent \textbf{Cross-Task Complementary Fusion (CTCF).}
To obtain more reliable ensemble logits $E_i$ for window-based regions, we propose the CTCF block, as shown in Fig.~\ref{framework}. It adaptively fuses SAM $\textbf{F}_{T}$'s instance mask outputs $I_i$ with TA $\textbf{F}_{TA}$'s semantic segmentation outputs $S_i$.
The objective of the fusion is to assign the highest-confidence semantic label to each instance mask based on the logits $S_i$ from TA (See Fig.~\ref{framework}).
Different from directly finding the most frequent or area-dominant semantic label~\cite{chen2023semantic,chen2023segment}, we define distinct area proportion thresholds for masks of different sizes and adaptively select the most reliable label.
Specifically, for each instance mask $K_i^m$ from SAM, we assess the instance mask area. For larger and smaller instance masks, we set the threshold $\theta$ to a smaller value to facilitate obtaining the most frequently occurring label. For medium-sized masks, we increase $\theta$ to a larger value to ensure the acquisition of a more accurate semantic label.
Then, we identify the top three semantic labels in descending order of quantity within the corresponding area in TA's predictions.
If the label coverage rate \textit{(lcr}) of the most prevalent semantic label $Y_{max}$ exceeds the $\theta $, we directly assign the most prevalent label as the semantic label for the current instance mask $K_i^m$. If the coverage rate of the most prevalent label falls below $\theta $, we delve into the three semantic labels and calculate their Shannon entropy (SE) using the logits $S_i$ from TA. The label with the minimum entropy is chosen as the highest-confidence semantic label $Y_{instance}$ for the $K_i^m$. The formulation is as follows:
\begin{small}
\begin{equation}
\setlength{\abovedisplayskip}{3pt}
\setlength{\belowdisplayskip}{3pt}
Y_{instance}=\left\{
\begin{array}{rcl}
 Y_{max},& & {lcr(Y_{max})\geq \theta},\\
\\
Y_{argmin\{SE(Y_a)\}}, & & {0 <{lcr(Y_a)}< \theta},
\end{array} \right.
\end{equation}
\end{small}
where $a$ belongs to the top three semantic labels occupying the instance mask. The detailed pseudo-code can be found in the \textit{Supplmat}.
Through the CTCF block, DAR produces high-quality ensemble logits by adaptively merging the predictions based on windows from both SAM and TA.

Due to the fusion process potentially resulting in varying prediction confidences among masks, we obtain a weight map ($M_i$) with spatial dimensions identical to $W_i$ in the fusion process. Specifically, we assign higher weight (1) to masks with higher overlap between instance masks and semantic logits and lower weight (0) to masks that require SE for label assignment.
The ensemble logits and weight maps obtained by the CTCF block can assist the TA and the student in achieving better supervision for patch predictions.

\subsection{Multi-level Knowledge Adaptation Module}
\label{MKA}
After addressing the distortion problem with our DAR module, we then propose the MKA module to learn a compact student module with TA $\textbf{F}_{TA}$'s output logits $S_i$  and the ensemble logits $E_i$ from the DAR module for multi-level, multi-scale (whole-image and patch) knowledge adaptation.
To effectively transfer knowledge from the whole-image scale prediction of the TA $\textbf{F}_{TA}$ to the student model $\textbf{F}_{S}$, we concatenate predictions of non-overlapping patches to generate the entire ERP semantic prediction map $S_{TA}^{w}$.
Therefore, when the entire image is directly fed to the student model $\textbf{F}_{S}$, resulting in the prediction map $P_{S}^{w}$. We use the Cross-Entropy (CE) loss (Eq.(\ref{eq:ce(TA,S)})) to guide the student in aligning its predictions with $S_{TA}^{w}$ at whole-image scale:.
\begin{small}
\begin{equation}
\setlength{\abovedisplayskip}{5pt}
\setlength{\belowdisplayskip}{5pt}
  \mathcal{L}_{ce}^{TA,S}= \mathcal{L}_{CE}(S_{TA}^{w},P_{S}^{w}).
  \label{eq:ce(TA,S)}
\end{equation}
\end{small}

For the obtained ensemble logits $E_i$ based on the patch $W_i$, we use another CE loss to guide the student's prediction logits at the corresponding window position $P_{S}^{i}$ to mimic $E_i$.
As the CTCF block returns the weight maps $M_i$ corresponding to the higher confidence masks based on the fusion mechanism, we first combine weight maps with ensemble logits to obtain higher confidence masks $E_H^i$. Then we perform knowledge adaptation simultaneously using ensemble logits $E_i$ and higher confidence masks $E_H^i$.
Therefore, the loss for patches knowledge from $E_i$ and $E_H^i$ transfer to the student $\textbf{F}_{S}$ and TA $\textbf{F}_{TA}$ can be formulated as:
\begin{small}
\begin{equation}
  \mathcal{L}_{CE}^{T,S}=\sum_{k=1}^{H\times W}(\mathcal{L}_{CE}(P_{i}^k,E_i^k)+\lambda M_i\mathcal{L}_{CE}(P_{i}^k,E_i^k)),
  \label{eq:ce(TA,S)}
\end{equation}
\end{small}
\begin{small}
\begin{equation}
  \mathcal{L}_{CE}^{T,TA}=\sum_{k=1}^{H\times W}(\mathcal{L}_{CE}(S_{i}^k,E_i^k)+\lambda M_i\mathcal{L}_{CE}(S_{i}^k,E_i^k)),
  \label{eq:traning loss of S}
\end{equation}
\end{small}
where $\lambda$ is the hyper-parameter.
By considering higher confidence masks $E_H^i$, we can enhance the learning of both the student and TA towards the reliable ensemble logits during the update process. This helps mitigate the impact of noisy labels generated by SE in lower confidence masks on both the student and TA.
The patches supervision not only assists the student in refining the segmentation of large objects but also contributes to improved recognition of smaller objects.

Additionally, we utilize the refined boundary map $B_{ref}^i$ obtained from DAR to supervise the student's boundary map ($B_S^i$) in the corresponding overlapping area $O_i$, enhancing the student's awareness of boundaries.
\begin{small}
\begin{equation}
\setlength{\abovedisplayskip}{3pt}
\setlength{\belowdisplayskip}{3pt}
\begin{aligned}
\mathcal{L}_{bd}^{T,S}=\sum_{k = 1}^{H\times W}\frac{|B_{ref}^i-B_S^i|}{C_o}.
\label{eq:boundary loss of T and S}
\end{aligned}
\end{equation}
\end{small}
Therefore, the total loss employed for the student $\textbf{F}_{S}$ comprises three components:
\begin{small}
\begin{equation}
\setlength{\abovedisplayskip}{5pt}
\setlength{\belowdisplayskip}{5pt}
  \mathcal{L}_{student}= \mathcal{L}_{CE}^{TA,S} +  \mathcal{L}_{CE}^{T,S} + \mathcal{L}_{bd}^{T,S}.
  \label{eq:traning loss of St}
\end{equation}
\end{small}
The total loss for the TA $\textbf{F}_{TA}$ is formulated as follows:
\begin{small}
\begin{equation}
\setlength{\abovedisplayskip}{5pt}
\setlength{\belowdisplayskip}{5pt}
  \mathcal{L}_{TA}= \mathcal{L}_{CE}^{T,TA} +  \mathcal{L}_{CC} + \mathcal{L}_{bd}^{T,TA}.
  \label{eq:traning loss of TA}
\end{equation}
\end{small}

\section{Experiments}

\begin{figure*}[t!]
    \centering
    \includegraphics[width=\textwidth]{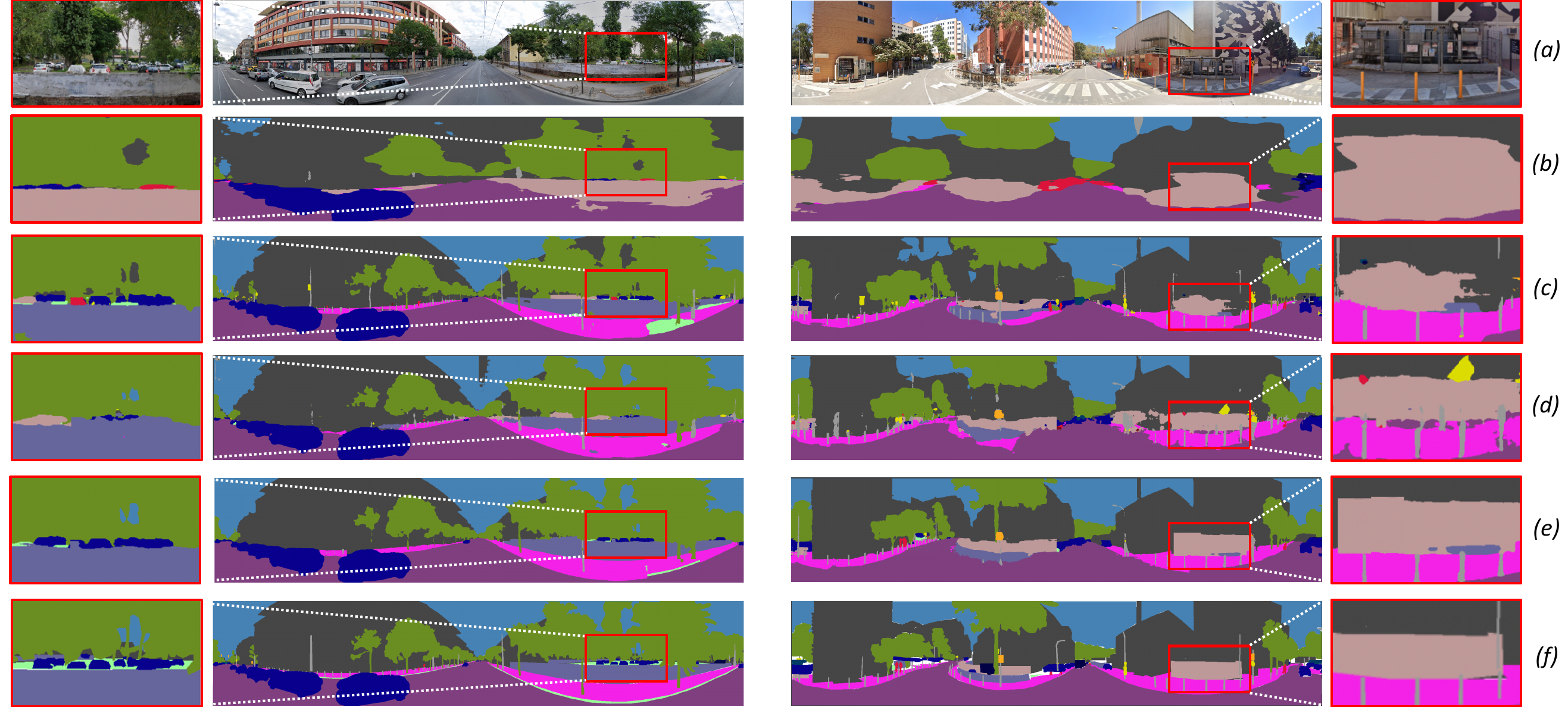}
    \vspace{-16pt}
    \caption{Example visualization results from the DensePASS test set: (a) Input panorama image, (b) Segformer-B5~\cite{xie2021segformer} without sliding window sampling, (c) DPPASS-S~\cite{zheng2023both}, (d) DATR-S~\cite{zheng2023look}, (e) GoodSAM-S, (f) Ground truth.}
    \vspace{-6pt}
    \label{visFigure}
\end{figure*}

\begin{table*}[t!]
\centering
\setlength{\tabcolsep}{2pt}
\resizebox{\textwidth}{!}{
\begin{tabular}{l|c|c|ccccccccccccccccccc}
\toprule
Method & P. (M) & mIoU  & Road  & S.W. & Build. & Wall  & Fence & Pole & Tr.L. & Tr.S. & Veget. & Terr. & Sky & Person & Rider & Car & Truck & Bus & Train & M.C. & B.C. \\ \midrule
ERFNet~\cite{romera2017erfnet}&- &16.65 & 63.59 & 18.22    & 47.01   & 9.45 & 12.79 & 17.00  & 8.12  & 6.41 & 34.24 & 10.15 & 18.43 & 4.96  & 2.31  & 46.03 & 3.19  & 0.59  & 0.00  &8.30  & 5.55   \\
PASS(ERFNet)~\cite{yang2019pass}      &- & 23.66 & 67.84 & 28.75    & 59.69    & 19.96 & 29.41 & 8.26  & 4.54          & 8.07         & 64.96      & 13.75   & 33.50 & 12.87  & 3.17  & 48.26 & 2.17  & 0.82  & 0.29  & 23.76      & 19.46   \\
Omni-sup(ECANet)~\cite{yang2020omnisupervised}   &- & 43.02 & 81.60 & 19.46    & 81.00    & 32.02 & 39.47 & 25.54 & 3.85          & 17.38        & 79.01      & 39.75   & \underline{94.60} & 46.39  & 12.98 & 81.96 & 49.25 & 28.29 & 0.00  & 55.36      & 29.47   \\
P2PDA(Adversarial)~\cite{zhang2021transfer} &- & 41.99 & 70.21 & 30.24    & 78.44    & 26.72 & 28.44 & 14.02 & 11.67         & 5.79         & 68.54      & 38.20   & 85.97 & 28.14  & 0.00  & 70.36 & 60.49 & 38.90 & 77.80 & 39.85      & 24.02   \\
PCS~\cite{yue2021prototypical}&25.56 & 53.83 & 78.10 & 46.24 & 86.24  & 30.33 &45.78 & 34.04  & 22.74  & 13.00 &\underline{79.98} & 33.07 & 93.44 & 47.69  & 22.53  & 79.20 & 61.59  & 67.09  & 83.26  & 58.68  & 39.80   \\
Trans4PASS-T~\cite{zhang2022bending} &13.95 & 53.18 & 78.13 & 41.19    & 85.93    & 29.88 & 37.02  & 32.54 & 21.59         & 18.94        & 78.67      &\textbf{45.20}   &93.88 & 48.54  & 16.91 & 79.58 & 65.33 & 55.76 & 84.63 & 59.05      & 37.61   \\
Trans4PASS-S~\cite{zhang2022bending} &24.98 & 55.22 & 78.38 & 41.58    & 86.48 & 31.54 & 45.54  & 33.92 & 22.96 & 18.27 & 79.40 & 41.07 & 93.82 & 48.85  & 23.36 &81.02 &67.31 &\textbf{69.53} & 86.13 & 60.85 & 39.09   \\
DPPASS-T~\cite{zheng2023both}  &14.0 &55.30 &78.74 &46.29 &87.47 &\textbf{48.62} &40.47 &35.38 &24.97 &17.39 &79.23 &40.85 &93.49 &52.09 &\underline{29.40} &79.19 &58.73 &47.24 &86.48 &66.60 &38.11 \\
DPPASS-S~\cite{zheng2023both} &25.4 &56.28 &78.99 &48.14 &87.63 &42.12 &44.85 &34.95 &27.38 &19.21 &78.55 &\underline{43.08} &92.83 &\textbf{55.99} &29.10 &80.95 &61.42 &55.68 &79.70 &\underline{70.42} &38.40         \\
DATR-M~\cite{zheng2023look}  &4.64 &52.90 &78.71 &48.43 &86.92 &34.92 &43.90 &33.43 &22.39 &17.15 &78.55 &28.38 &93.72 &52.08 &13.24 &77.92 &56.73 &59.53 &93.98 &51.52 &34.06 \\
DATR-T~\cite{zheng2023look} &14.72 &54.60 &79.43 &49.70 &87.39 &37.91 &44.85 &35.06 &25.16 &19.33 &78.73 &25.75 &93.60 &53.52 &20.20 &78.07 &60.43 &55.82 &91.11 &67.03 &34.32         \\
DATR-S~\cite{zheng2023look} &25.76 &56.81 &\underline{80.63} &51.77 &87.80 &44.94 &43.73 &\textbf{37.23} &25.66 &\textbf{21.00} &78.61 &26.68 &93.77 &54.62 &\textbf{29.50} &80.03 &67.35 &63.75 &87.67 &67.57 &37.10 \\ \midrule
GoodSAM-M(ours) &3.7 &55.93 &79.57 &51.04 &86.24 &43.42 &44.86 &30.92 &26.60 &\underline{20.62} &77.79 &25.43 &92.99 &53.77 &25.84 &82.01 &70.94 &62.29 &91.93 &58.24 &38.25\\
GoodSAM-T(ours) &14.0 &\underline{58.21} &80.06 &\textbf{53.29} &\underline{89.75} &44.91 &\underline{46.98} &31.13 &\underline{27.81} &19.83 &79.58 &25.72 &93.81 &\underline{55.44} &26.99 &\underline{84.54} &\underline{73.07} &68.41 &\underline{93.99} &67.36 &\underline{43.39}\\
GoodSAM-S(ours) &25.4 &\textbf{60.56} &\textbf{80.98} &\underline{52.96} &\textbf{93.22} &\underline{48.17} &\textbf{51.28} &\underline{33.51} &\textbf{28.09} &20.15 &\textbf{81.64} &30.97 &\textbf{95.21} &55.13 &29.01 &\textbf{87.89} &\textbf{75.28} &\underline{69.37} &\textbf{94.98} &\textbf{73.28} &\textbf{49.64}\\
\bottomrule
 \end{tabular}}
\vspace{-4pt}
\caption{Per-class results of the SOTA panoramic image semantic segmentation methods on DensePASS test set. (P.: Param.)}
\vspace{-8pt}
\label{perclass}
\end{table*}

\subsection{Datasets and Implementation Details}

\textbf{Dataset.} We leverage two benchmark datasets WildPASS~\cite{yang2021context} and DensePASS~\cite{ma2021densepass} to assess the segmentation performance of the GoodSAM.
The resolution of images in both datasets utilized is 400$\times$2048.

\noindent \textbf{Implementation details.} We train the proposed framework with PyTorch in 4 NVIDIA A6000 GPUs.
We keep SAM frozen during our experiments and utilize it solely for providing instance masks and boundary information.
For the TA and student models, we opt for the fine-tuned Segformer~\cite{xie2021segformer} series, encompassing B0-B5 variants, which come in six different sizes and exhibit varying performance levels in 2D image semantic segmentation.
We set the window size as $400\times 512$, with stride 256. The hyper-parameter $\alpha$ is set to 0.3. For medium-sized masks, the area range is from 100 to 1000. The thresholds $\theta$ for masks of different areas are set to 0.5 and 0.7. The hyper-parameters of weight for reliable masks are set to 0.2.
\textit{More details can be found in the supplmat.}

\subsection{Comparisons with Existing Works}
We first compare GoodSAM with previous panoramic semantic segmentation methods, including ERFNet~\cite{romera2017erfnet}, PASS~\cite{yang2019pass}, Omni-sup~\cite{yang2020omnisupervised}, P2PDA~\cite{zhang2021transfer}, PCS~\cite{yue2021prototypical}, Trans4PASS~\cite{zhang2022bending}, DPPASS~\cite{zheng2023both}, and DATR~\cite{zheng2023look}, on the DensePASS dataset. As shown in Tab.~\ref{perclass}, our GoodSAM-M, GoodSAM-T, and GoodSAM-S consistently exhibit superior performance at their respective parameter levels. Specifically, Our GoodSAM-S outperforms DATR-S, DPPASS-S, and Trans4PASS-S by \textbf{3.75}\% IoU, \textbf{4.28}\% IoU, and \textbf{5.34}\% IoU, respectively, which yields SOTA performance. Additionally, GoodSAM-M achieves a competitive mIoU of \textbf{55.93}\% mIoU with just \textbf{3.7} million parameters, which is comparable to DATR-T and surpasses Trans4PASS-T.
For the segmentation performance of each class, our GoodSAM-S outperforms the others in the majority of classes, including `building' (+\textbf{5.42}\% IoU), `fence' (+\textbf{5.5}\% IoU), and nearly all types of transportation (\textit{e.g.}, `car' with +\textbf{5.93}\% IoU).
Fig.~\ref{visFigure} shows visual comparisons of GoodSAM-S with other methods on the Densepass evaluation set. This showcases that our GoodSAM-S can generate distortion-aware and boundary-enhanced logits under the supervision of SAM and TA.

Tab.~\ref{wildpass} presents the experimental results on the WildPASS validation dataset. Our GoodSAM outperforms the existing methods based on Segformer B1 and B2 backbones. Specifically, our GoodSAM-T outperforms DPPASS-T by \textbf{2.04}\% mIoU on the Segformer-B1 backbone, and our GoodSAM-S surpasses DPPASS-S by \textbf{1.65}\% mIoU on the Segformer-B2 backbone. This indicates that, even with an expanded number of evaluation images, our GoodSAM consistently achieves superior performance.

\begin{table}[t!]
\centering
\setlength{\tabcolsep}{10pt}
\resizebox{\linewidth}{!}{
\begin{tabular}{ccc}
\toprule
Method                      & Backbone           & mIoU(\%) \\ \midrule
\multirow{2}*{Source domain Supervised} & Segformer-B1~\cite{xie2021segformer}     & 47.90         \\
                            & Segformer-B2~\cite{xie2021segformer}       &  54.11       \\ \midrule
Trans4PASS-T~\cite{zhang2022bending}                & Segformer-B1  &54.67          \\
Trans4PASS-S~\cite{zhang2022bending}                & Segformer-B2  &62.91          \\
DPPASS-T~\cite{zheng2023both}                & Segformer-B1       & 60.38       \\
DPPASS-S~\cite{zheng2023both}                  & Segformer-B2       & 63.53         \\
GoodSAM-M(ours)                  & Segformer-B0      & 58.65         \\
GoodSAM-T(ours)                  & Segformer-B1       & 62.42         \\
GoodSAM-s(ours)                  & Segformer-B2       & \textbf{65.18 }        \\
 \bottomrule
\end{tabular}}
\vspace{-8pt}
\caption{Experimental results of the SOTA panoramic image semantic segmentation methods on WildPASS test set.}
\vspace{-8pt}
\label{wildpass}
\end{table}

\begin{figure}[t]
    \centering
    \includegraphics[width=\columnwidth]{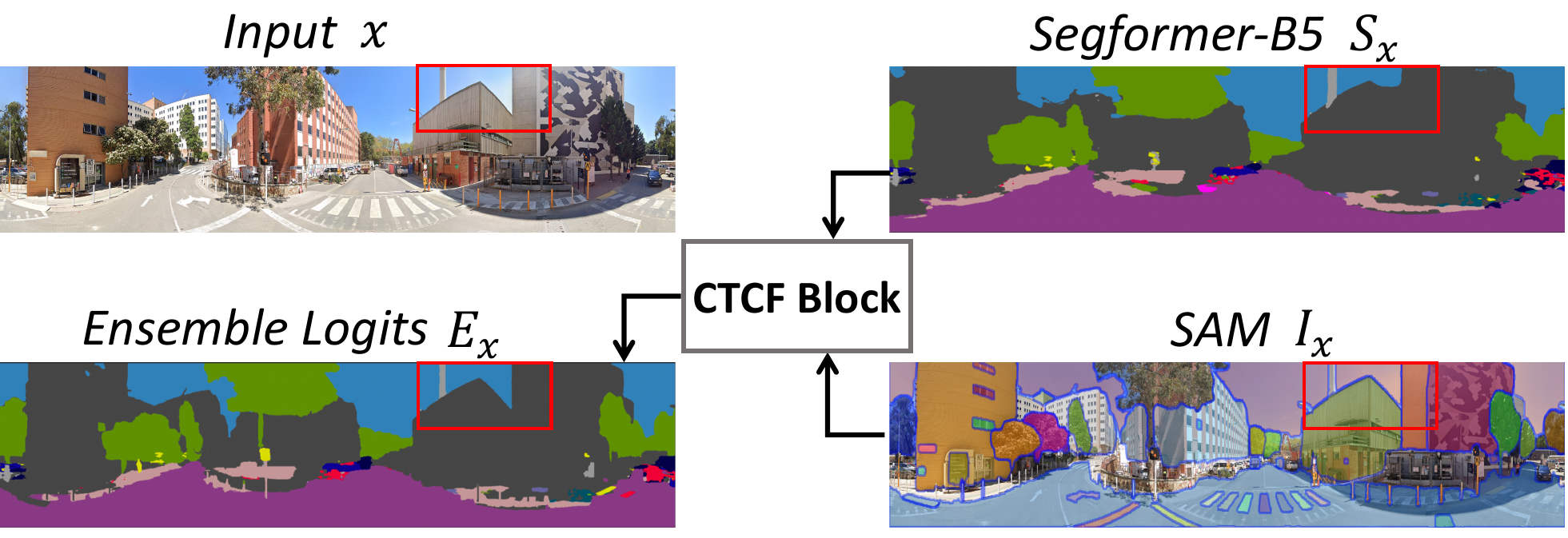}
    \vspace{-10pt}
    \caption{Effectiveness of the CTCF block.
    }
    \vspace{-10pt}
    \label{fig:CTCF}
\end{figure}

\subsection{Ablation Studies and Analysis}
\subsubsection{Effectiveness of DAR module}
Tab.~\ref{tab:Learning Scheme in Training and Testing} illustrates the effectiveness of each component of our framework. We choose Segformer-B5 as the TA model and Segformer-B0 as the student model.

\noindent \textbf{1) Effectiveness of sliding window sampling.} The performance comparison before and after employing sliding window sampling for the DAR module reveals a significant performance gap. This is attributed to Segformer being trained on 2D images, and when directly applied to ERP images with larger FoV characteristics, the model shows a substantial decline in performance. Specifically, the performance gap is nearly twice as significant without the sliding window strategy (B5: \textbf{27.62}\% mIoU vs. \textbf{53.86}\% mIoU, B0: \textbf{15.88}\% mIoU vs. \textbf{39.35}\% mIoU).

\noindent \textbf{2) Effectiveness of CTCF.} We evaluate the CTCF block based on the non-overlapping sliding window strategy. We propose CTCF to obtain patch ensemble logits and then rebuild the entire ERP prediction map at corresponding positions. As shown in Tab.~\ref{FusionW}, the mIoU of the ensemble logits currently stands at \textbf{55.88}\% mIoU, surpassing TA's performance by \textbf{2.02}\% mIoU. Furthermore, as we utilize $\mathcal{L}_{ce}^{T,TA}$ to update TA based on patch ensemble logits, we observe a continuous improvement(+\textbf{5.1}\% mIoU) in ensemble logits performance alongside the enhancement of TA's performance.  Fig.~\ref{fig:CTCF} demonstrates the visual differences between the segmentation map produced by Segformer-B5 as TA and the ensemble logits obtained through CTCF block,  highlighting the effectiveness of our CTCF block.


\noindent \textbf{3) Effectiveness of consistency constraint.}
we now ablate the consistency constraint based on the overlapping sliding window. Introducing the consistency constraint $\mathcal{L}_{CC}$ to implicitly mitigate the inconsistencies in overlapping regions between adjacent windows caused by distortion significantly improves the TA's performance(+\textbf{1.18}\% mIoU).

\begin{figure}[t]
    \centering
    \includegraphics[width=0.9\columnwidth]{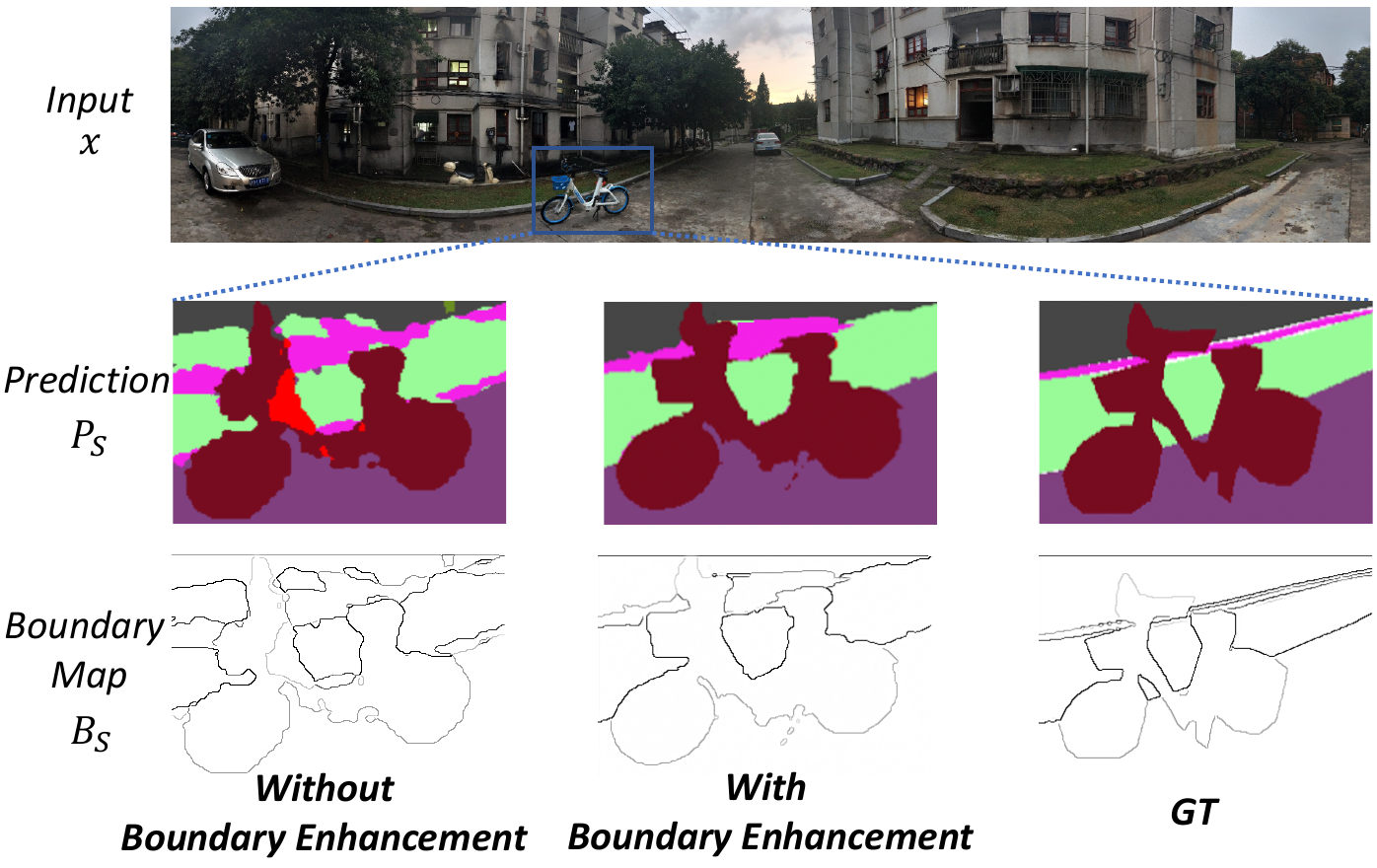}
    \vspace{-5pt}
    \caption{Effectiveness of the boundary enhancement block.
    }
    \vspace{-12pt}
    \label{fig:boundary}
\end{figure}

\noindent \textbf{4) Effectiveness of boundary enhancement.} We propose the boundary enhancement block to leverage the boundary information provided by SAM, enhancing TA's ability to predict boundary pixels. As revealed by Tab.~\ref{tab:Learning Scheme in Training and Testing}, the introduced boundary-enhanced loss boosts TA's performance by \textbf{2.42}\% mIoU (from\textbf{ 60.07}\% mIoU to \textbf{62.49}\% mIoU). The right part of visualization results in Fig.~\ref{visFigure} demonstrates our superior boundary segmentation performance for `fence' and `sidewalk' compared to previous methods. Meanwhile, Fig.~\ref{fig:boundary} also illustrates the impact of the boundary enhancement block on student predictions. These experiment results indicate that our boundary enhancement block explicitly assists TA and student in increasing awareness of boundaries and addressing distortion problem.

\vspace{-10pt}
\subsubsection{Effectiveness of MKA module}
For the MKA module, as shown in Tab.~\ref{tab:Learning Scheme in Training and Testing}, we observe that the performance of the student is only \textbf{15.88}\% mIoU when the entire ERP image is directly input. However, as we update the student using TA's logits, the performance increases to \textbf{50.90}\% mIoU with the improvement of TA's performance. Furthermore, by supervising the semantic logits of the student in the corresponding region using window-based ensemble logits, the performance of our student improves by \textbf{3.22}\% mIoU. Finally, by further constraining the student's boundary pixel prediction results with the refined boundary map obtained from DAR, our student's performance has been elevated to \textbf{55.93}\% mIoU. This indicates the  effectiveness of MKA in endowing GoodSAM with distortion-aware and boundary-aware capabilities.

\begin{table}[t]
\centering
\setlength{\tabcolsep}{2pt}
\resizebox{\linewidth}{!}{
\begin{tabular}{cc|ccc|ccc|cccc}
\midrule
\multicolumn{2}{c}{SAM and TA}& \multicolumn{3}{c}{TA} & \multicolumn{3}{c}{Student}& \multicolumn{3}{c}{mIoU}\\
\midrule
SW & CTCF & $\mathcal{L}_{ce}^{T,TA}$ & $\mathcal{L}_{CC}$ & $\mathcal{L}_{bd}^{T,TA}$ & $\mathcal{L}_{CE}^{TA,S}$ &$\mathcal{L}_{CE}^{T,S}$ & $\mathcal{L}_{bd}^{T,S}$ & E.S. & TA & Student\\
\midrule
- & -& -& -&- & -&-&- &-  &27.62 &15.88\\
$\surd$ & -& -& -&- & -&-&- &-  &53.86 &15.88\\
$\surd$ & $\surd$ & -& - & -&-&- & - &55.88 &53.86 &15.88\\
\midrule
$\surd$ & $\surd$ & $\surd$& - & -&$\surd$&- & - &60.68 &58.89 &49.88\\
$\surd$ & $\surd$ &$\surd$ &$\surd$ &-&$\surd$ &- &-&-  &60.07 &50.07\\
$\surd$ & $\surd$ &$\surd$ & $\surd$ & $\surd$&$\surd$ & -&-&-  &62.49 &50.90 \\
\midrule
$\surd$ & $\surd$ &$\surd$ & $\surd$ & $\surd$&$\surd$ & $\surd$&- &- &62.49 &54.12 \\
$\surd$ & $\surd$ &$\surd$ & $\surd$ & $\surd$&$\surd$& $\surd$&$\surd$&- &62.49 &55.93 \\
\bottomrule
\end{tabular}}
\vspace{-8pt}
\caption{Ablation of loss functions during the training process.
E.S. represents ensemble logits. }
\vspace{-8pt}
\label{tab:Learning Scheme in Training and Testing}
\end{table}

\vspace{-8pt}
\subsubsection{Other Analysis}
\textbf{Hyper-parameters analysis in sliding window sampling.}
Tab.~\ref{SW size} presents the effects of four different window sizes and step sizes with Segformer-B2. We observe that TA exhibits the best performance (\textbf{53.86}\%) when the window size is 400$\times$512. Meanwhile, the ensemble logits obtained after fusion with SAM achieve the highest performance (\textbf{55.88}\%). For the sliding window stride, we find that as the stride decreases, the overlapping area between adjacent windows increases, resulting in an increase in the number of windows and training overhead for TA. Therefore, based on Tab.~\ref{SW size}, due to constraints on training resources, when the stride is set to 256, TA achieves the highest performance(\textbf{60.07}\%) under the constraints of $\mathcal{L}_{CE}^{T,TA}$ and $\mathcal{L}_{CC}$.

\noindent \textbf{Analysis about CTCF module.}
We introduce the CTCF block to adaptively combine the outputs of SAM and TA, obtaining patch ensemble logits. Now, we evaluate our CTCF by comparing it with current methods that combine instance masks and semantic logits~\cite{chen2023semantic,chen2023segment}. SSA~\cite{chen2023semantic} assigns the label of the instance mask as the label that appears most frequently in the corresponding region of the semantic map. The fusion mechanism in SEPL~\cite{chen2023segment} is similar to SSA. They analyze each instance mask, selecting the label if it occupies more than half of the area in the corresponding region of the semantic map or if the distribution is almost covered by the instance mask.
However, we analyze the instance masks of different sizes during the experiment. We observe that the comparable coverage rate between the top two labels often occurs in medium-sized masks, which increases the risk of errors if the label with the highest rate is directly chosen. Therefore, we set different thresholds for different area sizes and incorporate SE to identify the highest-confidence label, ensuring an adaptive fusion of instance masks and semantic logits. Tab.~\ref{FusionW} illustrates that our fusion mechanism enhances the robustness and accuracy of the fusion process compared to the other two methods.

\begin{table}[t]
\centering
\setlength{\tabcolsep}{7pt}
\resizebox{0.48\textwidth}{!}{
\begin{tabular}{ccccccc}
\toprule
Window Size  &  400$\times$1024 & 400$\times$512  & 400$\times$256  & 400$\times$128  \\
\midrule
mIoU  &52.45 &\textbf{53.86} &53.29 &52.79   \\    \midrule
Strides  & 512  & 400 & 320  & 256    \\
\midrule
mIoU  &58.89 &59.68 &59.73 &\textbf{60.07}    \\ \bottomrule
\end{tabular}}
\vspace{-8pt}
\caption{Ablation about threshold window size and the step size of the sliding windows.}
\vspace{-8pt}
\label{SW size}
\end{table}

\begin{figure}[t]
    \centering
    \includegraphics[width=\columnwidth]{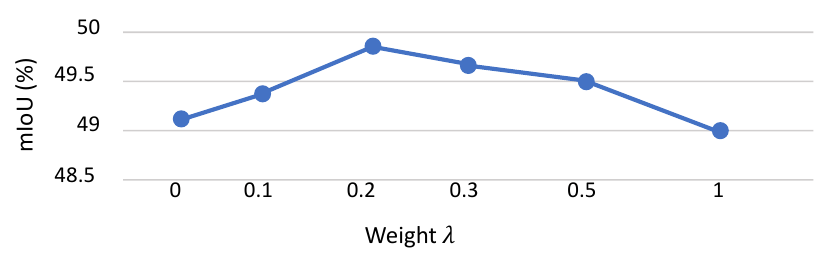}
    \vspace{-20pt}
    \caption{Ablation study for $\lambda$ of knowledge tranfer from ensemble logits to our student.
    }
    \vspace{-10pt}
    \label{fig:weight}
\end{figure}

\noindent \textbf{Hyper-parameter analysis in the loss functions.}
When transferring knowledge from window-based ensemble logits to the TA and student, we assign higher weights to the instance masks with high confidence for the CTCF block. Fig.~\ref{fig:weight} shows that with $\lambda$ set to 0.2, our GoodSAM can more efficiently extract correct knowledge from ensemble logits.

\noindent \textbf{The impact of TA selection.}
We assess the impact of different TAs on the training of the student. As shown in Tab.~\ref{FusionW}, we observe that as the performance of TA improves, our student's performance also increases. Specifically, when TA is Segformer-B5, our student achieves a performance gain of +\textbf{3.4}\% mIoU compared to TA being Segformer-B3. This improvement is attributed to Segformer-B5's ability to provide more comprehensive and accurate semantic logits.

\begin{table}[t]
\centering
\setlength{\tabcolsep}{4.8pt}
\resizebox{0.48\textwidth}{!}{
\begin{tabular}{ccccccccc}
\toprule
& & $B_0$  & $B_1$  & $B_2$  & $B_3$  & $B_4$ & $B_5$ \\ \midrule
&SSA~\cite{chen2023semantic}  &39.28 & 40.28 &47.93 & 49.13 & 52.19 & 54.32    \\
\multirow{2}*{Fusion}&SEPL~\cite{chen2023segment}  &39.68 &40.75 & 48.98 &50.37   &52.53 &54.58  \\
&CTCF &40.68 &41.75 & 49.7 &52.89   &54.13 &55.88 \\
\midrule
Student&$B_2$ &- &- &- &57.16 &59.02 &60.56&\\
\bottomrule
\end{tabular}}
\vspace{-8pt}
\caption{Ablation of fusion ways and different TA model.}
\vspace{-8pt}
\label{FusionW}
\end{table}


\section{Conclusion and Future Work}
In this paper, we designed a comprehensive framework for lightweight panorama semantic segmentation, which leverages the assistance of SAM and TA. By addressing the distortion and large FoV problems in panoramic images and bridging the capacity gap between SAM and the student model, our GoodSAM produced distortion-aware and boundary-enhanced logits, surpassing SOTA UDA methods across various model parameter levels.

\noindent \textbf{Future work:}
It would be worthwhile to fine-tune SAM to realize a foundational segmentation model suitable for panoramic images.
Additionally, we plan to investigate methods for distilling SAM's zero-shot capabilities into our compact segmentation model.

\noindent \textbf{Acknowledgment:} This paper was supported by the Guangzhou 2024 Applied Basic Research Project (Co-funded by Municipal Schools (Institutes) and Enterprises) Fund under Grant No. 2024A04J4072.
\clearpage

{
    \small
    \bibliographystyle{ieeenat_fullname}
    \bibliography{main}
}


\end{document}